\documentclass[sigconf,authorversion]{acmart}
\usepackage{booktabs} 
\usepackage{diagbox}
\usepackage{color}

\usepackage{subcaption}
\usepackage[lined,boxed,commentsnumbered]{algorithm2e}
\usepackage{epstopdf}
\usepackage{graphicx}
\usepackage{multirow}

\newcommand{\showDOI}[1]{\unskip}
\newcommand{\shownote}[1]{\unskip}
\newcommand{\showURL}[1]{\unskip}

\copyrightyear{2017} 
\acmYear{2017} 
\setcopyright{acmcopyright}
\acmConference{GECCO '17}{July 15-19, 2017}{Berlin, Germany}\acmPrice{15.00}\acmDOI{http://dx.doi.org/10.1145/3071178.3071232}
\acmISBN{978-1-4503-4920-8/17/07}

\begin{document}
\title[Investigation of Environmental Influence on Adaptation Mechanisms]{An Investigation of Environmental Influence on the Benefits of Adaptation Mechanisms in Evolutionary Swarm Robotics
}

\author{Andreas Steyven}
\orcid{0000-0001-9570-3697}
\affiliation{
  \institution{Edinburgh Napier University}
 \streetaddress{10 Colinton Road}
  \city{Edinburgh}
  \country{Scotland, UK}
}
\email{a.steyven@napier.ac.uk}

\author{Emma Hart}
\affiliation{
  \institution{Edinburgh Napier University}
 \streetaddress{10 Colinton Road}
  \city{Edinburgh}
  \state{Scotland, UK}
}
\email{e.hart@napier.ac.uk}

\author{Ben Paechter}
\affiliation{
  \institution{Edinburgh Napier University}
 \streetaddress{10 Colinton Road}
  \city{Edinburgh}
  \country{Scotland, UK}
}
\email{b.paechter@napier.ac.uk}


\begin{abstract}
A robotic swarm that is required to operate for long periods in a potentially unknown environment can use both evolution and individual learning methods in order to adapt. However, the role played by the environment in influencing the effectiveness of each type of learning is not well understood. In this paper, we address this question by analysing the performance of a swarm in a range of simulated, dynamic environments where a distributed evolutionary algorithm for evolving a controller is augmented with a number of different individual learning mechanisms. The learning mechanisms themselves are defined by parameters which can be either fixed or inherited. We conduct experiments in a range of dynamic environments whose characteristics are varied so as to present different opportunities for learning. Results enable us to map environmental characteristics to the most effective learning algorithm.





\end{abstract}


\maketitle

\section{Introduction}
\label{sec:introduction}

Recent advances in technology are driving novel research in swarm
robotics, envisioning future applications in which swarms might be
sent to remote or hazardous environments and in which they will need
to survive over long periods of time.  As these environments will be
unknown to the designer {\em a priori} and are potentially dynamic,
the swarm must be able to continuously adapt its behaviour to ensure
it both maintains sufficient energy to survive, and to successfully
perform tasks.

The importance of being able to adapt over time has been a subject of
research within Evolutionary Robotics for some time \cite{Walker2006}.  
Adaptation
often takes one or all of three
forms: evolutionary, individual and social learning. In {\em evolutionary}
adaptation,
information encoded on the genome adapts through selection
and reproductive operators over many generations. In {\em individual}
learning, a robot can adapt its own behaviour during the course of its
lifetime, for example, updating weight values in a neural network
controller. Finally in {\em social} learning, robots can exchange
information during a lifetime.

The relative benefits of mixing the different types of 
adaptation
have
been studied both in simulation \cite{Haasdijk2013a,Ellefsen2013} and hardware
\cite{Heinerman2015a,Heinerman2015,Heinerman2016,Gomes2016}. Typically, experiments are conducted in single environment
related to a specific task, therefore the role of the
environment in influencing the result is not made explicit. An
exception is recent work from Haasdjik \cite{Haasdijk2015} who explicitly studied
the effect of combining conflicting environmental and task
requirements in a simulated system. This showed that high selective
pressure exerted by a task can outweigh any selective pressure from
the environment. However, an arbitrary environment was defined to
conduct experiments in, leaving open the question of whether the same
effects would be observed in a different environment.


The goal of this paper is to investigate the interplay between
evolution, individual learning and environment characteristics. We consider a swarm which undergoes distributed evolution
of a neural-network based controller and is
augmented with an individual learning mechanism: this modifies the
information gleaned from the environment and fed to the controller
over the lifetime of a robot. Specifically, we consider a swarm
operating in an environment which is unknown {\em a priori} and which
robots must learn relative values of positive and negative energy
tokens. Each environment contains $n$ positive and $n$ negative energy
tokens. Positive tokens increase the robot's energy by $v$ units of
energy, while negative ones reduce it by a fixed amount. As $n,v$
vary, each environment presents different opportunities for learning
in that there are a small number of high value tokens, or a large
number of low value tokens. In addition, tokens change their nature
across 'seasons', i.e. tokens of a specific colour switch value from
negative to positive on a cyclical basis.  This forces the swarm to
have to re-learn the effect of any given colour of token every
season. Various 
settings
for individual learning are investigated in
which the learning mechanism is either fixed or has components that
can be simultaneously evolved. The following questions are
investigated:

\begin{itemize}
\item How do the parameters of the environment (token count, token
  value) influence the effectiveness of different individual learning
  settings?

\item How does the rate of change of a given environment influence
  the effectiveness of  individual learning mechanisms?

\item How does the nature of the individual learning mechanism
  influence performance in different environments?

\end{itemize}

We augment a distributed evolutionary algorithm previously described
in  \cite{Hart2015} with mechanisms for individual
learning in order to conduct experiments.  Note that the goal is not
to propose a novel method of either individual learning or evolutionary adaptation
but to explore the relationship between the environment and
value of different types of adaptation.





\section{Related Work}


A reasonable body of research exists in relation to combining learning
and evolution,
and factors that
influence this relationship \cite{Mayley1996,Nolfi1999,Haasdijk2008}. 
  The relationship of the two methods in a swarm environment in
which it is necessary to simultaneously learn behaviours which enable
reproduction in addition to task performance is less well studied
however. Haasdjik {\em et al} propose a framework for
evolution, individual and social learning in collective systems, and
consider the interaction of evolution and individual learning in
which the latter is achieved by {\em reinforcement learning}
\cite{Sutton1998}. Their experiments show that in a collective
system, it is possible for learning to counteract evolution.  A {\em
  hiding-effect} can occur in which individual learning acts to mask
the ill-adapted nature of non-optimal agents and is therefore
counter-productive. Although a number of environments were investigated
which essentially modified the reward system, all environments were
static, and the relationship of the learning framework to specific
parameterisations of the environmental features was not examined.

A dynamically changing reward system was investigated in
\cite{Bredeche2010} who proposed mEDEA, a completely
distributed evolutionary algorithm for open-ended evolution. Here,
efficient adaptation in a changing environment was demonstrated using
a set up that switched 
phases: in the {\em free-ride} phase,
there is no cost to movement therefore a robot only needs to meet a
single other robot to pass on its genome, while in the alternating
phase the robot is required to harvest energy in order to move and
therefore creating opportunties for passing on its genome. Haasdijk
{\em et al} \cite{Haasdijk2013} extended mEDEA to add explicit task-selection in
the MONEE framework \cite{Noskov2013}. In \cite{Haasdijk2015} they examine in
more detail the relative selection pressures induced by task
performance and survival in different environments, finding that task
performance is optimised even if it reduces the lifetime of robots
(and therefore their ability to reproduce).  Heinermann {\em et al}
investigate the relationship between evolution, individual and
social learning in real swarm \cite{Heinerman2015a,Heinerman2015,Heinerman2016}. Here, the
evolutionary part focuses on evolving a suitable sensory layout, while
the individual learning runs an evolution strategy to learn the
network weights during the robot lifetime. Learnt weight vectors are
broadcast to other robots during the social learning phase. The main
focus of this work was to investigate the impact of social
learning. Individual learning is {\em required} to learn a controller
and hence cannot be omitted.

In contrast to the above, we consider scenarios in which individual
learning has the potential to improve evolved behaviours, but is not
essential. We investigate the relative benefits of evolution and
individual learning using a variety of learning mechanisms and in a
range of environments with different features. The goal is to
specifically relate the roles of evolution and individual learning
performance to features of the environment.

\section{Overview}

A swarm operates in an open environment in which there are two types
of coloured tokens: driving over one colour increases the robots
energy while the other decreases it. Robots should learn to avoid the
negative token. However, a ``seasonal'' change is imposed where the
value of the token is reversed, i.e.  red becomes positive and blue
negative or vice versa. A robot must thus adapt any previously evolved
behaviour. All robots in the swarm evolve a neural network that
controls their behaviour through a distributed evolutionary algorithm \cite{Hart2015}
In addition, they can exploit an
individual learning mechanism which can potentially learn the {\em
  current} value of a given colour of token. This information modifies
an input to the evolved neural network. We investigate a number of
types of individual learning in which some components of the learning
mechanism can be either heritable, fixed or absent.


Experiments are conducted using the Roborobo simulator
\cite{Bredeche2013}. 
The robots have 8 ray-sensors distributed around
the body and detect proximity to the nearest object and its type. 
Each
robot is controlled by an evolved Elman recurrent neural network
(RNN). The network has 16
sensory inputs and 2 motor outputs (translational and rotational speeds). 
The 16 inputs comprise of two information of each of the 8 ray-sensors, proximity and whether or not this object is an energy token.
Although the colour/type of the object is also detected by the robot, it is not fed into the RNN as an input, but only used in the 
adaptation mechanism\footnote{the information cannot be encoded directly to the network
  without {\em a priori} knowledge of the number of potential colours}.


\subsection{mEDEA}
Using the inputs and outputs just described, an RNN with 1 hidden
layer containing 16 nodes is evolved by a distributed evolutionary
algorithm \cite{Hart2015}. This algorithm is an extension of $mEDEA$
\cite{Bredeche2010}, and incorporates a selection mechanism based on
relative fitness. In brief,
for a fixed period, robots move according to their control algorithm,
broadcasting their genome that is received and stored by any robot
within range.  At the end of this period, a robot uses
fitness-proportionate selection to choose a genome from
its list of collected genomes according to a relative fitness value,
and applies a variation operator.  This takes the form of a Gaussian
random mutation operator, inspired from Evolution
Strategies. Pseudo-code is given in Algorithm
\ref{alg:rf-medea_rf_pseudo_code}.

\begin{algorithm}
  load($currentGenome = randomInitialisedGenome$)\;
  \While{iteration $\le$ maxIterations}{
    \If{$\textrm{hasGenome()}$}{
      \eIf{lifetime $\le$ maxLifetime \& energy $> 0$}{
        move()\;
        \If{neighbourhood.$\textrm{isNotEmpty()}$}{
                $rf = $calculateRelativeFitness()\tcp*[r]{eq.\ref{eq:rf-fi}}\label{alg:rf-medea_rf_pseudo_code-calc-rf}
                broadcast($currentGenome$,$rf$)\;\label{alg:rf-medea_rf_pseudo_code-broadcast}
              }
      }{
            remove($currentGenome$)\;
          }
    }
    $genomeList$.addIfUnique($receivedGenomes$)\;\label{alg:rf-medea_rf_pseudo_code-receive}
    \If{genomeList.$\textrm{size()}$ $> 0$}{
      $genome =$ select$_{roulette-wheel}$($genomeList$)\;\label{alg:rf-medea_rf_pseudo_code-select}
          load($currentGenome =$ applyVariation($genome$))\;
          $genomeList$.empty()\;
          $lifetime = 0$\;
    }
  }
  \caption{
    \label{alg:rf-medea_rf_pseudo_code} Pseudo code of the adapted version of the mEDEA algorithm with relative fitness mEDEA$_{rf}$ as introduced in \cite{Hart2015} used with roulette-wheel as explicit selection mechanism}
\end{algorithm}

Each robot estimates its fitness in terms of its ability to survive
based on the balance between energy lost and energy gained, denoted
($\delta_E$): this term is initialised to 0 at $t=0$ (when the current
genome was activated) and is decreased according an energy-model
described below that accounts for both movement and the cost of
communicating for evolution, and increased by $E_{token}$ if it crosses an
energy token.  Given $\delta_E$, a robot calculates a fitness value
which is relative to those robots in the neighbourhood of range $r$.
according to equation \ref{eq:rf-fi}, where $f'_i$ is the relative
fitness of robot $i$ at time $t$, $mean_{sub_i}$ is the mean
$\delta_E$ of the robots within the subpopulation defined by all
robots in range $r$ of robot $i$, and $sd_{sub_i}$ is the standard
deviation of the $\delta_E$ of the subpopulation.

\begin{equation}
  f'_i(t)  = \frac{\delta_i(t) - mean_{sub_i}(t)}{sd_{sub_i}(t)}
\label{eq:rf-fi}
\end{equation}

There is a fixed cost to living of $0.5$ units per timestep, regardless of whether the robot moves or not.
A robot moving consumes an amount of energy that is related to its rotational speed $v_{\text{rot}}$, translational speed $v_{\text{trans}}$, and their respective maximum values $v_{\text{rot-max}}$ and $v_{\text{trans-max}}$

\begin{equation}
\label{eq:env-energy_motion}
    E_\text{step}
        = 0.5 + \left( \frac{v_{\text{rot}} }{ v_{\text{rot-max}} }
                + \frac{v_{\text{trans}} }{ v_{\text{trans-max}} } \right) / 4
\end{equation}

The amount of energy spent on communication $E_\text{com}$ is
calculated using equation \ref{eq:em-com-energy_total}, where $i$ and
$j$ are the number of genomes received and transmitted
respectively. The values $a_\textrm{rx}=0.0305$, $a_\textrm{tx}=0.01379$ and
$a_\textrm{tx-amp}=0.000614$ were determined based on the method described by
\cite{Steyven2015}; the reader is referred to this publication for a
description of their approach.
\begin{equation}
\label{eq:em-com-energy_total}
    E_{\text{com}}
        =   \sum_{k=0}^i 
                a_\text{rx}
            + \sum_{k=0}^j \left( 
                a_\text{tx} + b_\text{tx-amp} \times d^2 
                \right)
\end{equation}

Equation \ref{eq:rf-energy_change} shows the change in energy at each simulation step, where n is the number of tokens that have been collected in that step.
\begin{equation}
  E(t+1) = E(t) - E_\text{step} - E_\text{com} + (n_\text{token} \times E_\text{token})
  \label{eq:rf-energy_change}
\end{equation}

\subsection{Environment}
\label{sec:environment}
In Evolutionary Robotics, it is often unclear exactly how
parameterisation of a given environment might influence the emergence
of particular behaviours. Often, the focus of reported studies is on
algorithm performance, without serious consideration of how the choice
of environment may influence results.  This is particularly important
for an open-ended distributed algorithm such as $mEDEA$ in which
survival of robots is crucial for evolution to occur. To counter this,
Steyven {\em et al} \cite{Steyven2016a} recently proposed a technique by which preliminary
experimentation could be used to generate a surface-plot, highlighting
regions of the parameter space in which the
environment  provides the right balance between facilitating
survival and exerting sufficient pressure for new behaviours to
emerge. This enables a researcher to select appropriate settings for
experimentation.  For example, for a given task, on the one hand,
there will be regions in which the characteristics of the environment
are such that robots find survival to be trivial (e.g. food supplies
are unlimited and easy to find), and hence there is little pressure to
evolve specialised behaviours. On the other hand, environmental
characteristics which are harsh enough to cause individual robots to
die prematurely and therefore prevent any effective evolution are also identified.

\begin{figure}
  \centering
        \includegraphics[trim={0.2cm 0.7cm 1cm 2cm},clip,scale=0.36]{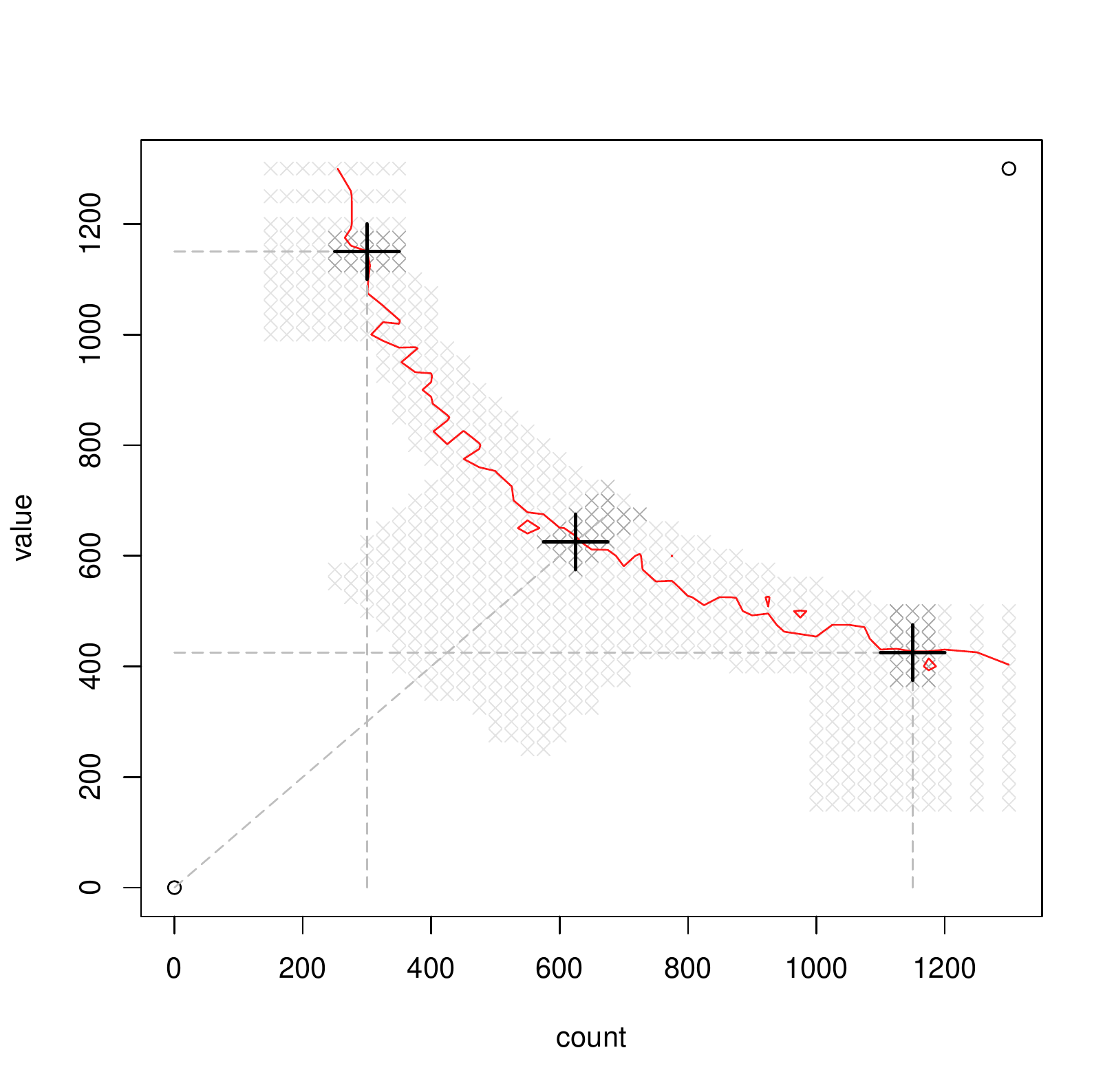}
    \caption{\label{fig:la-area-plot} Overview of newly created
      surface landscape. The red line shows the Neutral Line, the line where the surface plot crosses a plane drawn at delta-energy ($\delta_E$)=0.}
        
\end{figure}

Using the algorithm described above, we conducted experiments in an
environment parameterised by two variables: the {\em number} of energy
tokens available, and the {\em value} of the energy token. In each
environment tested, there are $n$ positive tokens with value $v$, and
$n$ negative tokens with value -400.
The delta-energy $\delta_E$,
i.e. difference between start and end energy is recorded for multiple
points in the parameter space, resulting in the plot shown in figure
\ref{fig:la-area-plot}. From this plot, we identify three points to
conduct experiments along the {\em energy neutral line}, i.e the
region in which the robot expends as much energy as it acquires. This
represents a region in which selection-pressure from the environment
to survive is neither too small or too large to mask the behaviours we
are interested in investigating.  The points identified are specified
in table \ref{tab:la-env_with-poison_configurations}.

\begin{table}*
  \centering
\caption{
      \label{tab:la-env_with-poison_configurations} Environmental
      configurations: description refers to the prevalence of energy
      tokens within the environment.}
    \begin{tabular}{ccc}
        \toprule
        \textit{Number of tokens} & \textit{Value per token} & \textit{Description} \\
        \midrule
        300   & 1150  & Scarce \\
        625   & 625   &  Balanced \\
        1150  & 425   & Abundant \\
        \bottomrule
    \end{tabular}
\end{table}

\section{Individual Learning}
The neural network described above has a set of 
 binary inputs (one for each sensor ray of the robot)
that denote the
presence (1) or absence (0) of a token (independent of its type).
Therefore, in an environment in which there are
multiple types of tokens, the only way for an individual to
distinguish between them is to pick up the token and observe the
change in energy.  If the environment in which the robot operates is
known {\em a priori}, then clearly, the neural network could be
designed in order to include relevant information about each token
type. However, if the environment is unknown, then the robot must
learn to adapt to the different types and values of tokens it may
encounter.

We use an adaptation mechanism which enables a robot to modify the
value input to the RNN corresponding to a token sensor: instead of simply having a binary
input, the robot uses a learned/evolved multiplier to adapt the token
input to a continuous value between $-1$ and $1$.

Each time a previously unseen type of token is encountered (detected
by a sensor ray or through consumption\footnote{The sensor rays of the
  robot are not evenly distributed around the robot body.  This can
  lead to the situation in which a robot drives over the token before
  any of the sensor rays detected it.}), a new multiplier is added to
the multiplier set.  As tokens are usually detected before they are
consumed, no information regarding a new token's value is known: the
robot therefore randomly initialises a value to associate with the
type ($x$) of detected token.  Following consumption, the resulting
change of energy is detected by the robot and its learning mechanism
can modify the corresponding multiplier value ($m_x$).

All multiplier values are adjusted every time a token is consumed
according to equation \ref{eq:la-multiplier_update}:

\begin{equation}
  m_x' = m_x + LS
          \times \bigg(LR-\frac{C_x}{C_{\textrm{total}}}\bigg)
          \times \Big(\frac{V_x}{V_{\textrm{max}} - V_{\textrm{min}}}\Big)
  \label{eq:la-multiplier_update}
\end{equation}

where $m_x$ is the current value for the multiplier for type $x$; $C_x
$is the number of tokens of type $x$ collected; $C_{\textrm{total}}$ is
the total number of all tokens collected; $V_x$ is the value of the
token that has just been consumed and is therefore now known to the
robot (being equivalent to the change in energy); $V_{\textrm{max}},
V_{\textrm{min}}$ define the minimum and maximum values of all tokens
encountered so far. $LR$ is a learning rate that controls the
magnitude of the change, and $LS$ is either $-1$ or $+1$ and simply
inverts the direction of change; this is required to adjust the
learning mechanism to the internal value notation of the neural
network and can be adapted via evolution.  The learning mechanism is
shown in Algorithm \ref{alg:la-multiplier_update}.


\begin{algorithm}
  \If{$token_x$ is unknown}{
      $multipliers$.add($token_x$)\;
  }
  \If{$token_x$ is consumed}{
    $tokenCounter_x$.update($token_x$)\;
    $totalTokenCount$.update()\;
    $tokenValue_x$.update($\delta_E(t)-\delta_E(t-1)$)\;
    $totalValueRange$.update()\;
    \For{$m_x$ in $multipliers$}{
        $m_x$.update()\tcp*[r]{eq. \ref{eq:la-multiplier_update}}
    }
  }
  \caption{
    \label{alg:la-multiplier_update} Pseudo code of the steps carried out to update all multipliers every time a token is encountered. }
\end{algorithm}




Three factors influence the learning mechanism: the initial value
assigned to a token $V_x$, the learning rate $LR$ and the associated
sign $LS$.  These factors can be randomly assigned, fixed to some
specific value, or can themselves be subject to evolution. Allowing
the learning sign to co-evolve enables the learning mechanism to
self-adapt to the internal value convention of the neural
network. Finally, enabling the robot to evolve an appropriate starting
value for each type of token based on its experience may speed up
learning in some circumstances. Even though token values change over
seasons, inheriting a good starting value may be beneficial, likely
dependent on the rate of change of the environment.

Table \ref{tab:learningalgs} defines four variants of the learning
algorithm that we investigate in conjunction with the three
environments described in section \ref{sec:environment}. Note that in no case is any
Lamarkian evolution used, i.e. although the multiplier starting values
are adapted over the course of a lifetime, they are {\em never}
written back to the genome and are therefore not inherited.


\begin{table}
\caption{\label{tab:learningalgs}Learning scenarios investigated showing heritability of information}
\begin{tabular}{cccc}
\toprule
 & Initial Value of Multiplier & LR & LS \\
\midrule
Baseline & 1 (all tokens) & none & n/a\\
IL & random & fixed & evolved \\
EVO & evolved & none &  n/a\\
EVO+IL & evolved & evolved & evolved \\
\bottomrule
\end{tabular}
\end{table}





\subsection{Experiments}

An experiment is defined by a tuple $<${\em environment, seasonal
  change rate, algorithm}$>$. Three environments (see section
\ref{sec:environment}) and three different rates of
seasonal change are investigated: 0 (no change, i.e. static
environment), every 5000 iterations, and every 15000
iterations. 
Note
that the maximum lifetime of a genome
before it is replaced is 2500
iterations, so every robot should 
go through
at least 
one evolutionary generation
during the
shorter (5000 iterations) season and at least 5 times in the 15000
season. In practice, as robots tend to die before their maximum
lifetime, more evolutionary cycles are likely to occur.

Four algorithms are investigated as detailed in table
\ref{tab:learningalgs}. Note that in the baseline experiments, all
tokens have a fixed multiplier of 1 and therefore the robots cannot
distinguish between tokens of different types.  Thus, in total 36
(=3x3x4) experiments are conducted. In each experiment, we record the
{\em totalTokenRatio} at the end of the season. This value is the
ratio of the number of collected token with positive value divided by
the sum of all collected token within that season.  A ratio of 0.5
shows that an equal amount of positive and negative token was
collected, below 0.5 more negative and above more positive token,
respectively.

\begin{table}
\centering
\caption{\label{tab:env-exp_params} Simulation and Experimental Parameters for all experiments}
\vspace{-2mm}

  \begin{tabular}[t]{ll}
    \toprule
    \multicolumn{2}{l}{\em Simulation parameters} \\
    \midrule
    Arena size  & 1024 px $\times$ 1024 px \\
    Max. robot lifetime   & 2500 iterations  \\
    Token re-spawn time  & 500 iterations \\
    Sensor range & 196 pixel \\
    Max. communication range $r_\mathrm{max}$ &  128 pixel\\
    \midrule
    \multicolumn{2}{l}{\em Experimental parameters} \\
    \midrule
    Number of independent runs  & 30 \\
    Number of robots  & 100 \\
    Max. iterations     & 1,000,000 \\
    Start energy & 500 \\
    \bottomrule
  \end{tabular}
\end{table}

Experimental and simulation parameters are given in table
\ref{tab:env-exp_params}. 
Parameters associated with the learning mechanism are given in
table \ref{tab:la-learning_params}.  The values for
$LR_\textrm{initial}$ and $LR_\textrm{max}$ where selected following
limited empirical exploration.  

The positive value of an energy
token is determined by the environment. In seasons when a token is
negative, the value is fixed -400 which is 80\% of a robot's initial
energy.

Following 30 runs of each experiment, statistical analysis was
conducted based on the method in \cite{Segura2016} using a significance level of 5\%. 
The distributions of two results were checked using a Shapiro-Wilk test.
If both followed a Gaussian distribution then Levene's test for homogeneity of variances was perfomed.
For equal variances the p-value was determined using an ANOVA test, otherwise using a Welch test.
A Kruskal-Wallis rank sum test was perfomed to determine the p-value if one of the results followed a non-Gaussian distribution.



   

  


\begin{table}
\begin{center}
\caption{\label{tab:la-learning_params} Learning parameter with their initial values and ranges in which they can change during runtime of the experiment.} 
\vspace{-2mm}
\begin{tabular}{lll}
  \toprule
  \textbf{Parameter}    & \textbf{Init. Value}  & \textbf{Value Range} \\
  \midrule
  Learning rate, $LR$
              & 1.02          & $[LR_\textrm{min},LR_\textrm{max}]$ \\
  Min. $LR$, $LR_\textrm{min}$ & 1             & fixed \\
  Max. $LR$, $LR_\textrm{max}$ & 1.5           & fixed \\
  Multiplier of type $x$,  $m_x$
                          & random          & $[-1,1]$ \\
  Learning sign, $LS$    & random          & $[-1,1]$ \\
  \bottomrule
\end{tabular}
\end{center}

\end{table}

\section{Results}
This section provides summarised results: detailed experimental data
is available as supplementary material.
Table \ref{tab:la-endvalues-by-count-and-season} shows the median totalTokenRatio for each of three
individual learning mechanisms (EVO, EVO+IL, IL) in each of the 3
environments and for each value of seasonal change. The values are
compared to the result from the baseline experiment each case, and
statistical significance is indicated in the table.

\begin{table*}
\caption{\label{tab:la-endvalues-by-count-and-season} Showing median of end values by seasonal change and Experiment for \emph{totalTokenRatio} over generation 199 to 200 (N:30).
$\downarrow$, $\leftrightarrow$, $\uparrow$ indicate whether the value
is lower, not different or higher respectively compared to the
baseline experiment. The number
of arrows corresponds to the magnitude level  of the effect size based
on a Vargha and Delaney A test. (1 = small, 2 = medium, 3 = large)}
\centering
\vspace{-2mm}
\setlength{\tabcolsep}{0.3em}
\begin{tabular}{|r|l||r|r|r||r|r|r||r|r|r|}
  \hline
  \multicolumn{2}{|r||}{\bf Experiment}
          & \multicolumn{3}{c||}{\bf Evo} 
            & \multicolumn{3}{c||}{\bf IL}
              & \multicolumn{3}{c|}{\bf Evo + IL} \\
  \cline{2-11}
          & {\bf count} 
          & {\bf 300}   & {\bf 625}   & {\bf 1150}  
            & {\bf 300}   & {\bf 625}   & {\bf 1150}  
              & {\bf 300}   & {\bf 625}   & {\bf 1150} \\
  \cline{2-11}
  {\bf Season}  & {\bf value} 
          & {\bf 1150}  & {\bf 625} & {\bf 425} 
            & {\bf 1150}  & {\bf 625} & {\bf 425} 
              & {\bf 1150}  & {\bf 625} & {\bf 425} \\
  \hline
  \hline
  \multicolumn{2}{|l||}{\bf 0}  
          & $\uparrow \uparrow \uparrow$ 0.5301 & $\uparrow \uparrow \uparrow$ 0.5411 & $\uparrow \uparrow \uparrow$ 0.5662 & $\leftrightarrow$ 0.5034 & $\leftrightarrow$ 0.5056 & $\downarrow$ 0.4997 & $\uparrow \uparrow \uparrow$ 0.5306 & $\uparrow \uparrow \uparrow$ 0.5388 & $\uparrow \uparrow \uparrow$ 0.5705 \\  
  \hline
  \multicolumn{2}{|l||}{\bf 5k}   
          & $\leftrightarrow$ 0.4995 & $\downarrow$ 0.4982 & $\downarrow$ 0.4989 & $\downarrow$ 0.5006 & $\downarrow \downarrow$ 0.4975 & $\leftrightarrow$ 0.5023 & $\leftrightarrow$ 0.5029 & $\uparrow \uparrow$ 0.5134 & $\uparrow \uparrow \uparrow$ 0.5191 \\ 
  \hline
  \multicolumn{2}{|l||}{\bf 15k}  
          & $\downarrow \downarrow$ 0.496 & $\downarrow \downarrow$ 0.495 & $\downarrow$ 0.4981 & $\downarrow$ 0.4973 & $\downarrow$ 0.4993 & $\downarrow$ 0.5011 & $\leftrightarrow$ 0.4981 & $\uparrow \uparrow$ 0.5136 & $\uparrow \uparrow \uparrow$ 0.5121 \\ 
  \hline
\end{tabular}
\end{table*}

The EVO method (which evolves multiplier values but has no adaption
during a lifetime) outperforms the baseline method in all three static
environments (season change = 0). Here, evolution is able to determine
appropriate values for each multiplier type. However, in the dynamic
environment, evolving the multiplier value is detrimental. In the
first season, evolution can find appropriate multiplier values
(particularly in a long season). However, as soon as the season
changes, these become irrelevant; if these values have spread
sufficiently through the population it may take considerable time for
evolution to reverse this change, while in the meantime, the robot
will continue to collect negative tokens.

The IL method (fixed learning rate and random initialisation of
values) never outperforms the baseline method in the static
environment, and is worse than the baseline in the dynamic
environments. The magnitude of the effect is highest in the seasonal
change=5000 environment for a balanced environment. 
It
  appears that the learning rate is not sufficient to adapt a randomly
  initialised multiplier to a suitable value while the randomness can
  actually bias the robot towards collecting a particular type. On
  average, this is worse than the baseline case in which the robot has
  equal preference for both types.



In contrast, with the exception of the two {\em dynamic and scarce}
environments, the EVO+IL method which evolves the $LR$, $LS$ and the multiplier values and also adjusts the latter during lifetime, a significant improvement is observed with
respect to the baseline method. In the {\em scarce} environments, the
robots have little information available to them to inform learning,
as there are few tokens. When the environment is changing rapidly this
is particularly detrimental. In the other environments, there are more
tokens to learn from. When this is coupled with the ability to both
evolve useful multiplier values {\em and} adapt them at a appropriate
rate, the swarm learns to adapt to the changing environments and
improves its behaviour in the static environment.

\subsection{Influence of environmental parameters}

Next, we examine the first question posed in section \ref{sec:introduction} in more
depth: {\em under what environmental conditions is augmenting
  evolution with an individual learning mechanism
  beneficial?}

Table \ref{tab:la-p-values-comparing-environments} provides a pairwise
comparison of environments for \emph{totalTokenRatio} obtained at the
end of each experiment.  In this table and subsequent ones, the symbols
$=$,
$<$,$>$ indicate whether the median values for
totalTokenRatio are not significantly different,
significantly smaller or larger respectively.
p-values below the significance level of 0.05 are written in bold.

Table \ref{tab:la-p-values-comparing-environments} clearly indicates that for the methods that
include an evolutionary component with the learning algorithm, then in
the static environment, $abundant$ $>$ $balanced$ $>$ $scarce$. In
contrast, when only a fixed individual learning mechanism is used with
no adaptation of learning rate, then the reverse appears true; the
token ratio is higher in the $balanced$ and $scare$ environments is
higher than in the $abundant$ environment, with no significant
difference between $balanced$ and $abundant$.

In the slow changing environment (15k), the general trend is that
$abundant$ $>$ $balanced$ $>$ $scarce$ for {\em all} three mechanisms.
In the rapidly changing environment, a mixed picture emerges. For the
EVO+IL mechanism, it is clear that $abundant$ $>$ $balanced$ $>$
$scarce$. For EVO, the $scarce$ environment does {\em not} provide
significantly different results to the other two, whereas for $IL$,
both $scarce$ and $balanced$ prove harder than $abundant$, but
$scarce$ outperforms $balanced$.


\begin{table*}
\centering
\caption{\label{tab:la-p-values-comparing-environments}  p-values of pairwise comparison of environments for \emph{totalTokenRatio} (row vs. column) over generation 199 to 200} 
\vspace{-2mm}
\begin{tabular}{|l|l|r||l|l||l|l||l|l|}
  \hline
  \multicolumn{3}{|r||}{\bf Experiment} &
          \multicolumn{2}{c||}{\bf Evo} & 
            \multicolumn{2}{c||}{\bf IL} & 
              \multicolumn{2}{c|}{\bf Evo + IL} \\
  \cline{2-9}
    & \multicolumn{2}{l||}{\bf count} 
          & {\bf 625}   & {\bf 1150}  
            & {\bf 625}   & {\bf 1150}  
              & {\bf 625}   & {\bf 1150} \\
  \cline{3-9}
  {\bf Season} &  & {\bf value} 
          & {\bf 625} & {\bf 425} 
            & {\bf 625} & {\bf 425} 
              & {\bf 625} & {\bf 425} \\
  \hline
  \hline
  {\bf 0}   & {\bf 300}   & {\bf 1150}  
          & $<$ \textbf{1.24e-07} & $<$ \textbf{3.02e-27}
            & $=$ 5.78e-01 & $>$ \textbf{7.33e-05}
              & $<$ \textbf{3.27e-02} & $<$ \textbf{1.44e-40} \\ 
  \cline{2-9}
        & {\bf 625}   & {\bf 625} 
          & & $<$ \textbf{9.37e-16}
            & & $>$ \textbf{7.87e-11}
              & & $<$ \textbf{1.33e-32} \\
  \hline
  \hline
  {\bf 5k}  & {\bf 300}   & {\bf 1150}  
          & $=$ 4.55e-01 & $=$ 3.08e-01 
            & $>$ \textbf{7.89e-05} & $<$ \textbf{1.99e-02} 
              & $<$ \textbf{3.87e-19} & $<$ \textbf{1.5e-32} \\ 
  \cline{2-9}
        & {\bf 625}   & {\bf 625} 
          & & $>$ \textbf{1.22e-03}
            & & $<$ \textbf{1.81e-17}
              & & $<$ \textbf{5.88e-04} \\
  \hline
  \hline
  {\bf 15k}   & {\bf 300}   & {\bf 1150}  
          & $<$ \textbf{4.78e-02} & $<$ \textbf{1.58e-04}
            & $=$ 6.51e-01 & $<$ \textbf{4.11e-04} 
              & $<$ \textbf{1.94e-16} & $<$ \textbf{9.56e-30} \\ 
  \cline{2-9}
        & {\bf 625}   & {\bf 625} 
          & & $<$ \textbf{1.09e-08}
            & & $<$ \textbf{1.44e-12}
              & & $=$ 2.61e-01 \\
  \hline
\end{tabular}

\end{table*}

\subsection{Influence of Environmental Change }
Table \ref{tab:la-p-values-comparing-seasons} illustrates how the rate of change of a given environment
influences the interaction between environmental parameters and
learning mechanisms. In 21/27 pairwise comparisons, statistically
significant results are observed.

In the $scarce$ environments, there is a general pattern that in terms
of rate of change, $static$ $>5k>15k$ for {\em all} mechanisms.  In
the $balanced$ environments, the same general pattern is observed,
with the exception that for the IL and EVO+IL mechanisms, no
statistical differences are noted between the 5k and 15k
environments. In the $abundant$ environments, we also note the same
general pattern as above, except that for IL,  the only significant
result shows that 5k$>$15k significant, while in contrast, for
EVO, 5k$<$15k.


\begin{table*}[ht]
\centering
\caption{\label{tab:la-p-values-comparing-seasons} Showing p-values of pairwise comparison of seasonal change for \emph{totalTokenRatio} (row vs. column) over generation 199 to 200} 
\vspace{-2mm}
\begin{tabular}{|l|l||l|l||l|l||l|l|}
  \hline
  \multicolumn{2}{|r||}{\bf Environment} &
          \multicolumn{2}{c||}{\bf count:300 value:1150} & 
            \multicolumn{2}{c||}{\bf count:625 value:625} & 
              \multicolumn{2}{c|}{\bf count:1150 value:425} \\
  \cline{2-8}
  {\bf Experiment}  & {\bf Season} 
          & {\bf 5k}  & {\bf 15k} 
            & {\bf 5k}  & {\bf 15k}
              & {\bf 5k}  & {\bf 15k} \\
  \hline
  \hline
  {\bf Evo}   & {\bf 0}
          & $>$ \textbf{9.09e-69} & $>$ \textbf{6.76e-55}
            & $>$ \textbf{1.2e-131} & $>$ \textbf{4.34e-99}
              & $>$ \textbf{6.91e-120} & $>$ \textbf{9.94e-88} \\ 
  \cline{2-8}
        & {\bf 5k}
          & & $>$ \textbf{1.89e-02}
            & & $>$ \textbf{1.67e-05}
              & & $<$ \textbf{6.04e-03} \\
  \hline
  \hline
  {\bf IL}  & {\bf 0}
          & $>$ \textbf{2.54e-05} & $>$ \textbf{6.93e-09}
            & $>$ \textbf{4.05e-27} & $>$ \textbf{4.43e-20}
              & $=$ 1.03e-01 & $=$ 7.08e-01 \\ 
  \cline{2-8}
        & {\bf 5k}
          & & $=$ 7.61e-02
            & & $=$ 8.51e-02
              & & $>$ \textbf{8.33e-03} \\
  \hline
  \hline
  {\bf Evo + IL}  & {\bf 0}
          & $>$ \textbf{2.82e-35} & $>$ \textbf{1.37e-31}
            & $>$ \textbf{1.76e-32} & $>$ \textbf{4.15e-32}
              & $>$ \textbf{3.41e-178} & $>$ \textbf{3.54e-118} \\ 
  \cline{2-8}
        & {\bf 5k}
          & & $>$ \textbf{1.7e-02}
            & & $=$ 4.38e-01
              & & $=$ 7.19e-01 \\
  \hline
\end{tabular}

\end{table*}

\subsection{Influence of learning mechanism}
Table \ref{tab:la-p-values-comparing-experiments} provides a pairwise
comparison of learning mechanisms within different environments. 22/27
comparisons are significant.

For the $scarce$ environment, general pattern that EVO+IL outperforms
the other two methods in 4/6 cases, with no statistical difference in
the other two cases. In the balanced environment, EVO+IL also clearly
dominates both EVO and IL.  EVO dominates IL in the static and 5k
experiments. Finally, in the $abundant$ environment, we again observe
the supremacy of EVO+IL, while IL dominates EVO in both of the dynamic
environments.


\begin{table*}
\centering
\caption{\label{tab:la-p-values-comparing-experiments} Showing
  p-values of pairwise comparison of  learning mechanism for \emph{totalTokenRatio} (row vs. column) over generation 199 to 200}
\vspace{-2mm}
\begin{tabular}{|l|l||l|l||l|l||l|l|}
  \hline
  \multicolumn{2}{|r||}{\bf Environment} &
          \multicolumn{2}{c||}{\bf count:300 value:1150} & 
            \multicolumn{2}{c||}{\bf count:625 value:625} & 
              \multicolumn{2}{c|}{\bf count:1150 value:425} \\
  \cline{2-8}
  {\bf Season}  & {\bf Experiment} 
          & {\bf IL}  & {\bf Evo + IL}  
            & {\bf IL}  & {\bf Evo + IL}
              & {\bf IL}  & {\bf Evo + IL} \\
  \hline
  \hline
  {\bf 0}   & {\bf Evo}
          & $>$ \textbf{6.04e-35} & $=$ 6.64e-01 
            & $>$ \textbf{8.51e-77} & $=$ 4.32e-01
              & $>$ \textbf{4.44e-82} & $<$ \textbf{1.5e-03} \\ 
  \cline{2-8}
        & {\bf IL}
          & & $<$ \textbf{3.6e-26}
            & & $<$ \textbf{6.66e-59}
              & & $<$ \textbf{6.7e-150} \\
  \hline
  \hline
  {\bf 5k}  & {\bf Evo}
          & $=$ 8.45e-01 & $<$ \textbf{3.18e-05}
            & $>$ \textbf{4.42e-06} & $<$ \textbf{3.36e-55}
              & $<$ \textbf{4.22e-15} & $<$ \textbf{9.94e-122} \\ 
  \cline{2-8}
        & {\bf IL}
          & & $<$ \textbf{1.27e-04}
            & & $<$ \textbf{3.6e-70}
              & & $<$ \textbf{2.9e-80} \\
  \hline
  \hline
  {\bf 15k}   & {\bf Evo}
          & $=$ 7.84e-02 & $<$ \textbf{8.55e-04}
            & $<$ \textbf{7.4e-03} & $<$ \textbf{1.09e-41}
              & $<$ \textbf{1.28e-08} & $<$ \textbf{3.11e-77} \\ 
  \cline{2-8}
        & {\bf IL}
          & & $=$ 5.27e-02
            & & $<$ \textbf{5.83e-42}
              & & $<$ \textbf{2.23e-72} \\
  \hline
\end{tabular}

\end{table*}

\subsection{Analysis}
\begin{figure}
\vspace{-2mm}
  \centering
  \includegraphics[page=1,trim={0.2cm 0.5cm 0cm 0cm},clip,scale=0.539]{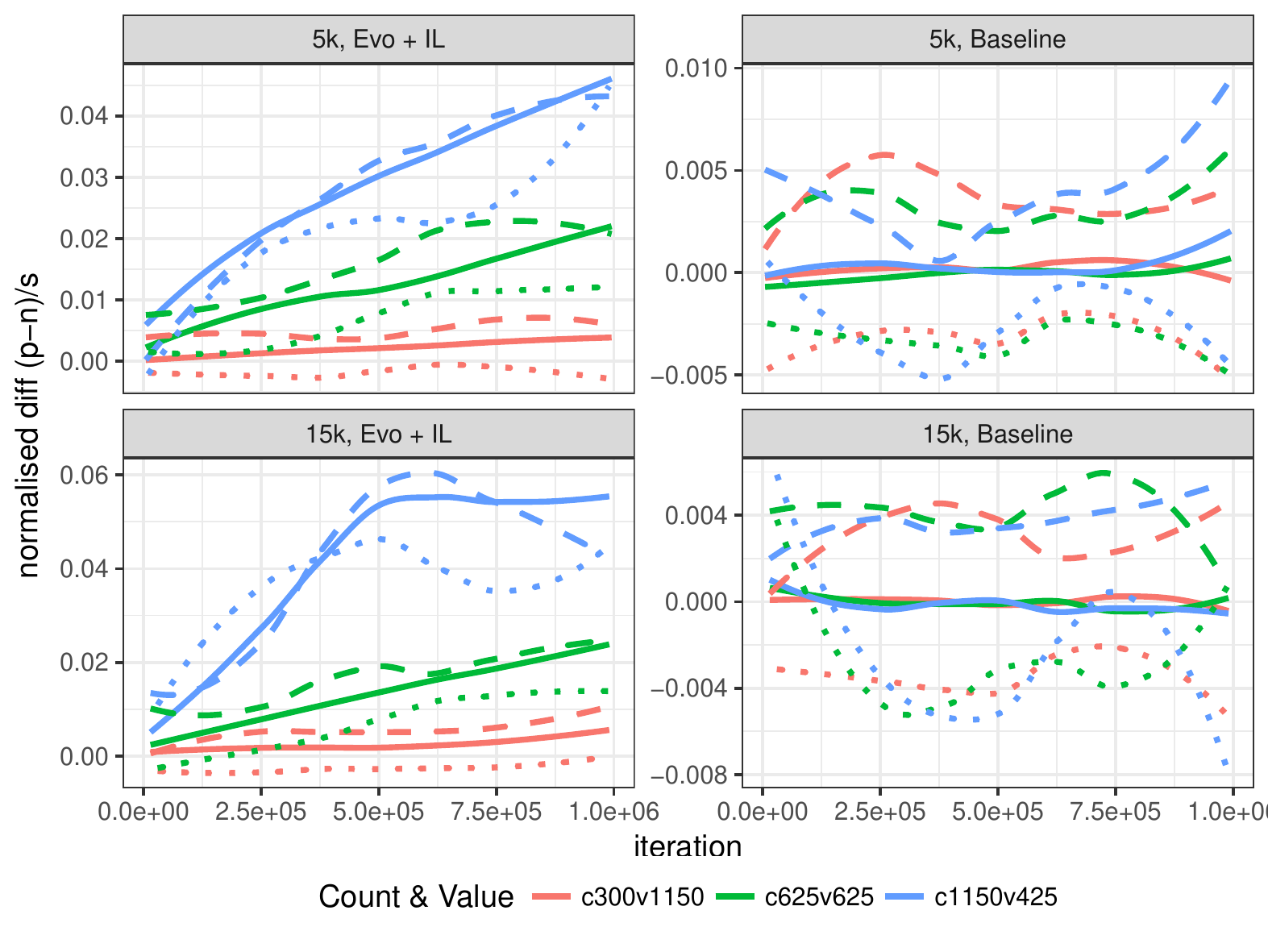}
  \caption{
      \label{fig:la-diff-comparison}  Normalised difference between
      positive and negative tokens collected. Solid line is value
      combined over all seasons, dashed = season 0, dotted = season 1  }
\end{figure}

The previous section showed that the EVO+IL clearly outperforms IL and
EVO in all parameterisations of the environment and for all rates of
change. We examine its behaviour more closely by plotting 
  the normalised difference between the number of positive tokens (p)
  and the number of negative tokens (n) collected per season over time
  (i.e. p-n).
    This is shown in figure \ref{fig:la-diff-comparison} for the (scarce,
  balanced, abundant) environments for the two cases in which the
  values of the tokens change dynamically with seasons.  The solid
  lines on the graph represent this value combined over both seasons,
  while the dashed and dotted lines represent the value in season 0
  and season 1 respectively. All lines are smoothed over the relevant
  points. The continuous improvement in this metric is clearly
  identified for EVO+IL, showing a  generally robust response to the changes in
  token value (i.e. an upward trend).  The $abundant$ environment proves most straightforward
  to learn in: having a large {\em quantity} of information of low-value 
outweighs the situation in which a small quantity of  high-value
information is available. In contrast, in the baseline experiment in
which $no$ information is available as to token value, the (p-n)
metric continuously cycles. In this case, the best that evolution can
do is learn a token-avoidance behaviour, as there is no means of
distinguishing between tokens.

\section{Conclusion}
We have investigated the performance of a number of adaptation
mechanisms that augment evolution
of a neural network
controller. Adaptation mechanisms that included heritable and fixed
components were analysed in three different environments in which both
the number of learning opportunities and the impact of the learning
opportunity varied.

We show that an adaptation mechanism in which all components evolve
and are heritable (EVO+IL) copes well in static and dynamic
environments, and is able to learn to distinguish between tokens of
different value. In dynamic environments, the greatest effect is
observed when the environment contains a large number of small
learning opportunities.  The fewer the learning opportunities, the
less effective the mechanism becomes, despite the fact that the
opportunities provide more energy and therefore more information to
the learning mechanism.

In contrast, the EVO and IL mechanisms  both prove to be detrimental in a changing
environment when compared to the baseline scenario. No clear pattern
emerges however in terms of the magnitude of the effect with respect
to the number of learning opportunities present. The IL method never
outperforms the baseline experiments, whereas EVO is beneficial only
in  a static environment.  In the latter case,  performance is
greatest in the environment with most tokens, and decreases as the
number of tokens decreases.

The results clearly demonstrate the interaction between the learning
mechanism and environmental parameters. This is of particular
relevance for distributed algorithms such as mEDEA in which 
environmental pressure influences reproductive abilities.  
The huge variety of behaviour that were displayed in different environments highlight how fundamental it is to not just select parameters at random, but to perform a more thorough analysis.
The emerging behaviour using a single set of algorithmic parameter varied from giving a massice advantage, to showing no difference, to even being counter productive.
Future work
will extend the analysis to other mechanisms for adding individual
learning and/or adaptation, as well as considering social learning,
recently demonstrated by \cite{Heinerman2015,Heinerman2016} to be effective in some scenarios.






\end{document}